\documentclass[10pt,twocolumn,letterpaper]{article}

\usepackage{cvpr}
\usepackage{times}
\usepackage{epsfig}
\usepackage{graphicx}
\usepackage{amsmath}
\usepackage{amssymb}
\graphicspath{ {images/} }

\usepackage[breaklinks=true,bookmarks=false,draft=false]{hyperref}

\cvprfinalcopy 

\def\cvprPaperID{****} 

\begin{document}

\title{Flow Based Self-supervised Pixel Embedding for Image Segmentation}

\author{
Bin Ma, Shubao Liu, Yingxuan Zhi, Qi Song\\
CuraCloud\\
999 3rd Ave Suite 700, Seattle, WA, USA, 98104\\
{\tt\small \{ binm, shubaol, derekz, song \} @curacloudcorp.com}
}

\maketitle

\begin{abstract}
   We propose a new self-supervised approach to image feature learning from motion cue. This new approach leverages recent advances in deep learning in two directions: 1) the success of training deep neural network in estimating optical flow in real data using synthetic flow data; and 2) emerging work in learning image features from motion cues, such as optical flow. Building on these, we demonstrate that image features can be learned in self-supervision by first training an optical flow estimator with synthetic flow data, and then learning image features from the estimated flows in real motion data. We demonstrate and evaluate this approach on an image segmentation task. Using the learned image feature representation, the network performs significantly better than the ones trained from scratch in few-shot segmentation tasks.
\end{abstract}

\section{Introduction}

In the past few years, Convolutional neural networks (CNN) have made tremendous progress in learning image features and solving various computer vision tasks~\cite{Krizhevsky2012,He2016,Ronneberger2015}. Most of the previous successes rely heavily on having access to large scale datasets with human annotated labels. However, the need to collect high quality human annotation significantly limits its applications in many applications, especially those in new domains, those that requires detailed annotation, such as image segmentation, where pixel-accurate annotation is required, and those that requires highly-skilled expert knowledge, such as medical applications.

Recently, there have been surging interest in learning image features without human provided labels, by exploring self-supervision signals that come naturally from data themselves. The key idea is to learn representations so that it can predict some attributes of data from other set of attributes. For example, it has been shown that image features can be learned by predicting next frame \cite{Lotter2016}, by predicting adjacent image patches \cite{Doersch2015}, by generating naturally looking images \cite{Goodfellow2014}, by recovering colors from grayscale images \cite{Larsson2017}.

In this paper, we study the problem of using motion as a self-supervision signal for feature learning. The basic idea is well explained in the one of the Gestalt principles: ``Elements that move in the same direction are perceived as more related than elements that are stationary or that move in different directions''. In video analysis, there are strong statistics that two pixels belong to the same object are more likely to move together than two pixels from different objects. Based on this, the motion cue provides a strong signal of two pixels having similar semantics.

On another thread, there have been recent works that show we can train deep neural network to perform certain vision tasks by training it on synthetic data, such as learning optical flow from synthetic flow \cite{Dosovitskiy2015,Sun2018,Hui2018}, feature point detection from synthetic images \cite{DeTone2017,Choy2016}, etc. These works show that these pixel-level prediction tasks are hard to label by hand, and are more suitable to train on synthetic data. Since the synthetic data are generated by composing natural images, the learned networks are shown to generalize well from synthetic to real data.

In this work, we show that by combining these two threads of recent work, we can learn good quality single image features by leveraging synthetic optical flow data. We demonstrate the quality of learned features on a medical image segmentation task, and show self-supervision with motion cue helps to learn features that generalize better than training from annotated data only, making it useful in settings where only limited annotation is available.

\section{Related work}
\subsection{Optical flow estimation}

Our work uses optical flow as a supervision signal to learn (motion-free) appearance-based pixel embedding. Optical flow, as a core computer vision concept, was traditionally formulated as an energy optimization problem introduced in \cite{Horn1981}. The optimization is usually done iteratively, which makes the flow estimation computationally expensive. Recently, Dosovitskiy et al.~\cite{Dosovitskiy2015} introduced a paradigm shift in flow estimation by showing the feasibility of directly estimating optical flow using convolutional neural network. In the new paradigm, the computational burden is transferred to the training step, while the estimation step can be more than two magnitudes faster. Compared with the traditional optimization based approach, the new learning based approach needs a large dataset of ground-truth optical flow for training. But it has been shown that synthetic flow data can be used to train CNN models that transfer well to real data, and generate good quality optical flow results. Synthetic flow data is usually generated by composing an object mask on a real image background with known motion as the ground-truth flow. Further, Sun et al.~\cite{Sun2018} introduced a new network architecture, called PWC-Net, and showed that it can outperform other flow estimation methods including traditional ones. Our work uses the PWC-Net to train from synthetic flow data and generate optical flow for real video data.

\subsection{Video segmentation}
A related problem is video segmentation, where the goal is to segment moving objects in a video sequence. Here the motion information is usually combined with appearance information together to segment moving objects, such as the ones based on traditional segmentation and learning techniques \cite{Stein2007}, and recent ones that is based on convolutional neural networks \cite{Jain2017}. In \cite{Cheng2017}, Cheng et al. even further combines the optical flow estimation together with object segmentation in one network architecture. In these works, the segmentation is inferred from information in multiple frames. This is related, but different from what we are doing here. Here we use the motion cue as a supervision signal to learn pixel embedding from one single image. After the learning, only a single image is necessary for the segmentation at the inference stage.

\subsection{Metric learning}
Our work is formulated as learning a metric embedding for each pixel so that their pairwise similarities matches the similarities measured in optical flow. Metric learning has been used in many computer vision problems. In face recognition, \cite{Schroff2015,Taigman2014,Chopra2005} learn metric embedding for each face so that faces from the same person have higher similarities than from other persons. In semantic learning, \cite{Fathi2017,DeBrabandere2017} learn metric embedding for each pixel, so that pixels from the same objects have higher similarities than others. Different from most of the previous metric embedding work, where the metric loss is defined on discrete values such as class labels, this work defines the metric loss using a continuous supervision signal estimated from inter-frame motion.

\subsection{Motion based self-supervision}

Motion as a supervision signal for training static image segmentation has also been studied before. \cite{Ross2009} uses background subtraction to get motion segmentation, which is then used to learn the image and shape properties in static images. More recently, \cite{Pathak2017} learns visual representation from motion segmentation, which is generated from a customized motion grouping method. \cite{Mahendran2018} learns pixel embedding such that it matches the optical flow directly without explicit motion segmentation. All the above methods use the traditional flow methods, such as the one described and implemented in \cite{Liu2009}. In this work, we show that recently developed deep learning based flow algorithms generate much sharper flows on the object boundary, which provides stronger signals and is more suitable in learning to segment static images. In the formulation of optical based loss function, we also properly handle the influence of flow magnitude.

\section{Method}
In this section, we describe our approach to learning static image features from synthetic flow data. More specifically, we learn feature representation for single image appearance --- a vector embedding for each pixel. Figure 1 shows the overall diagram of our approach. We first estimate the optical flow given two frames. From the estimated flow, we train an embedding based network to learn pixel embedding so that their pairwise similarities match the pairwise similarities derived from estimated optical flow.
\subsection{Flow consistency loss}
Given two pixels $p$ and $q$, let's denote the network embedding function as $\phi(p)$ and $\phi(q)$, denote their flows as $f_p$ and $f_q$. Then we measure the consistency between two pixel's appearance embedding and flow using the weighted sum of their cross entropy:
\begin{equation}
\mathcal{L}(\Theta) = \sum_p \| f_p \|_2 \mathcal{H}(P_f(\cdot \mid p),P_\phi(\cdot \mid p))
\end{equation}
Here, $f_p$ is the flow vector of pixel $p$; $\|f_p\|_2$ is the flow magnitude; $H(P_f(\cdot \mid p),P_\phi(\cdot \mid p))$ is the cross entropy between the optical flow defined transition probability, and the embedding defined transition probability at the pixel $p$. Here we weight the cross entropy by the flow magnitude at the pixel. The reason is that we are more confident about pixel grouping when the two pixels move fast together. As an extreme case, when no objects in the scene moves, the flow carries no information about pixel grouping. The cross entropy can be written as
\begin{equation}
    \mathcal{H}(P_f(\cdot \mid p),P_\phi(\cdot \mid p)) = \sum_q(P_f(q \mid p)\log(P_\phi(q \mid p)
\end{equation}

$P_\phi(q \mid p)$ is the probability of pixel $q$ being the best neighbor of pixel $p$ under the metric defined with the embedding function $\phi(\cdot)$.
It is defined as:
\begin{equation}
 P_\phi(q \mid p) = \frac{S_\phi(\phi_p,\phi_q)}{\sum_q S_\phi(\phi_p,\phi_q)}
\end{equation}
where $S_\phi(\phi_p,\phi_q)$ is the pixel embedding similarity between $p$ and $q$. It is measured using the cosine kernel:
\begin{equation}
 S_\phi(\phi_p,\phi_q) = \frac{\phi^T_p\phi_q}{\|\phi_p\|_2\|\phi_q\|_2}
\end{equation}

$P_f(q \mid p)$ is the probability of pixel $q$ being the best neighbor of pixel $p$ measured under the metric defined with optical flow $f(\cdot)$. 
Formally, it is defined as
\begin{equation}
 P_f(q \mid p) = \frac{S_f(f_q;f_p)}{\sum_q S_f(f_q;f_p)}
\end{equation}
where $S_f(f_q;f_p)$ is the similarity between pixel flow $p$ and $q$. It is measured using the Gaussian Radial Basis Function (RBF) kernel:
\begin{equation}
 S_f(f_q;f_p) = \exp\left(-\frac{\|f_p-f_q\|^2 / (\|f_p\|^2_2+\epsilon)}{2\sigma^2}\right)
\end{equation}
Here we use the relative flow magnitude in the above definition. This is because for the same pixel, its flow magnitude will change depending on the frame interval used to compute the optical flow. In practice, we put a small $\epsilon$ at the denominator to avoid numerical degeneracy. The value of $\epsilon$ depends on the optical flow's noise level. In the experiments reported in this paper, we empirically set $\epsilon = 1$ and $\sigma = 0.5$.

Here we handle flow magnitude differently from \cite{Mahendran2018}: the pixel flow similarities are defined based on relative flow magnitude; and the cross-entropy information are weighted based on the confidence of that information, which is measured by the magnitude of the flow.

\begin{figure}
    \center{\includegraphics[width=1.0\linewidth]{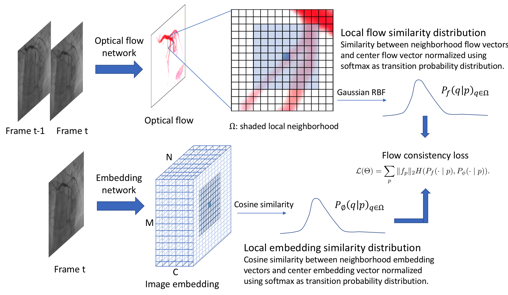}}
    \caption{Overall diagram of our embedding learning framework}
    \label{fig:diagram}
\end{figure}

\section{Experiments}
\subsection{Networks}
In this work, we utilize two networks, one for optical flow estimation, another one for pixel embedding. For optical flow estimation, we use the PWC-Net \cite{Sun2018}, which was shown to generate state-of-the-art results on several benchmarks by training on synthetic flow data, such as the Flying Chairs and Sintel. We use a PWC-Net pretrained on the Flying Chair dataset. For pixel embedding, we adopt the U-Net architecture and remove the last layer. But instead of training the U-Net with a classification loss supervised by the segmentation masks, we train it with a metric-learning loss supervised by optical flow estimated from PWC-Net.

\subsection{Dataset}
We evaluate the method on an image segmentation task using X-ray coronary angiography images. X-ray coronary angiography is a standard procedure in diagnosing coronary diseases in the catheter lab. This procedure involves injecting X-ray contrast medium into the coronary arteries via catheters. An X-ray camera films the blood flow to show the location and severity of artery narrowing over several heart beat cycles at normally 15 to 30 frames per second and saves the video in DICOM (Digital Imaging and Communications in Medicine) format. Each video file contains an image sequence of roughly 40 – 80 frames, each has 512*512 pixels, that dynamically captures the shape and state of the blood vessel. A common task is to perform blood vessel segmentation for subsequent Quantitative Coronary Analysis (QCA) and 3D blood vessel reconstruction from two or more stereo projection angles \cite{Kirbas2004}. While single-frame images offer strong signals on the appearance, the original DICOM data are temporal in nature and capture rich information about various object motion. The main motion patterns presented in the data are blood vessel motion due to heart beat, background organ movement due to breathing cycle, and sporadic global motion due to patient table movement. For this study, we use an X-ray angiography dataset consisting of 727 coronary angiography DICOM files that are randomly split into 582 training data and 145 testing data.

Our baseline method is to train a U-Net \cite{Ronneberger2015} model for blood vessel segmentation. For all the 727 DICOM files, we extract a middle frame from the image sequences and name it as frame 1 (i.e., f1), and generate its ground truth segmentation mask by first using Frangi filter \cite{Frangi1998} and then human refinement.

To demonstrate the proposed method, we also extract another frame 2 (i.e., f2) lagging by four frames from f1. The 4 frame interval is chosen to have large enough motion between two chosen frames. We obtain the corresponding optical flow map between the two frames respectively based on \cite{Liu2009} and \cite{Sun2018} pretrained on the Flying Chair dataset. As shown in Figure 2, we provide three sample data namely D1, D2 and D3 shown on three columns. On each column from top to bottom are segmentation mask of frame 1, frame 1, frame 2, and optical flow calculated by optical flow algorithm \cite{Liu2009} and PWC-Net \cite{Sun2018} pretrained on the Flying Chair dataset. As can be seen, optical flow highlights the area where there is motion, and similar objects tend to move together and thus have similar color as shown in the flow map. Flow result from \cite{Liu2009} appears to consist patches and is dilated, while flow based on PWC-Net is considerably sharper and shows vessel structure in greater details.

\begin{figure}
    \centering{\includegraphics[width=0.9\linewidth]{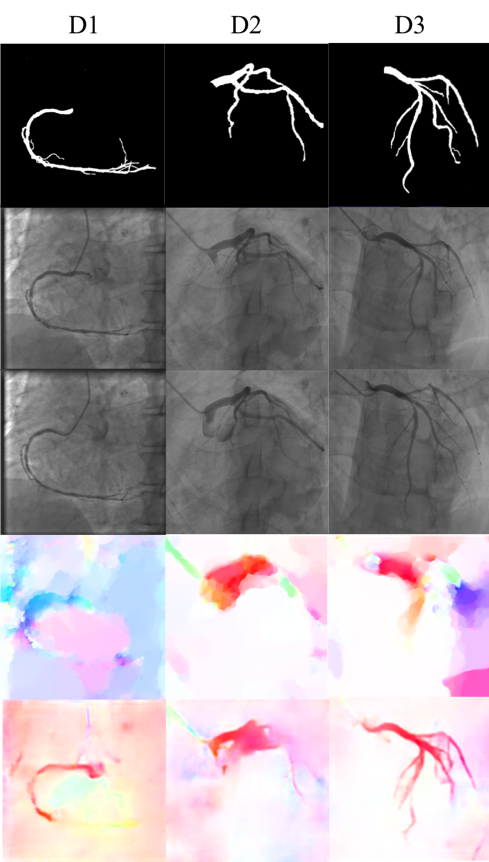}}
    \caption{Three typical data shown on three columns. At each column, from top to bottom: ground truth segmentation mask, frame 1, frame 2, optical flow predicted by \cite{Liu2009}, optical flow predicted by PWC-Net pretrained model \cite{Sun2018}.}
    \label{fig:b}
\end{figure}

\subsection{Training}
For our baseline segmentation model, we train a U-Net \cite{Ronneberger2015} using 1, 5, 10, 100 and 582 annotated training labels respectively. We consistently apply data augmentation (i.e., random horizontal flipping, random rotation between -20 and 20 degrees, random scaling between 0.8 and 1.2 and random shear transformation between -20 and 20 degrees) to alleviate overfitting. We modified original U-Net by adding batch normalization and padding before each convolution to maintain the height and width of the final embedding the same as the original image. We use binary cross entropy as the loss function and Adam optimizer with a fixed learning rate of 0.001. We threshold the predicted probability map with 0.5 and evaluate the segmentation performance with dice score. For our proposed self-supervised embedding method, we use the same U-Net except that the last layer is removed. The network outputs a 64 dimensional embedding vector for each pixel in the original image. We first trained the image embedding using flow signal. We normalized flow magnitude and use it as the sampling probability to randomly select 250 points without replacement from each image. For points that have no flow, they don't provide useful information for learning. For each point sampled, we train the model to learn the embedding similarity of the point and its 24 (i.e., 5*5-1) neighbors to be similar to the corresponding flow similarity. We trained the embedding model for 200 epochs and used Adam optimizer with a fixed learning rate of 0.001. We use a single Tesla V100 GPU with 16 GB memory and limit the batch size to 3.

After embedding training, we load the embedding model as a pretrained model and finetune it with annotated training labels. In order to have a fair comparison with the baseline U-Net model, we keep the training settings exactly the same.

\subsection{Results}
\begin{figure*}[htb]
    \center{\includegraphics[width=0.9\linewidth]{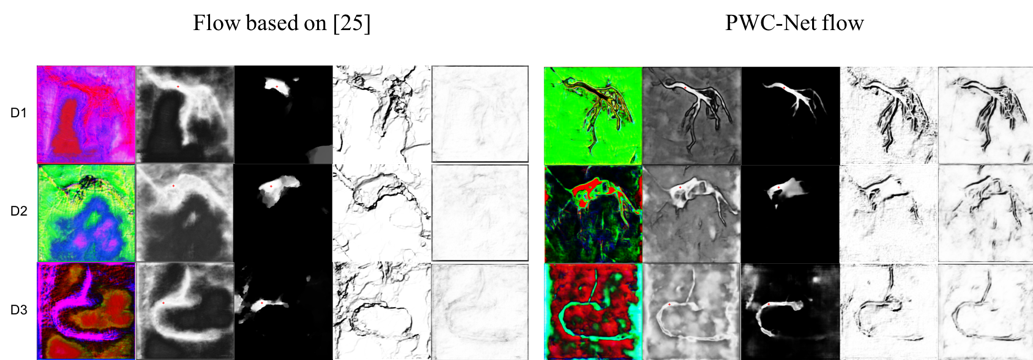}}
    \caption{For the same three data D1, D2, and D3, we visualize the learned embedding and similarity maps with [25] based flow supervision and PWC-Net based flow supervision on the left and right halves respectively. In each half of the figure, from left to right: random projection of image embedding, embedding similarity with an anchor vector shown as a red plus sign, flow similarity with an anchor vector, flow similarity between original and shifted flow, embedding similarity between original and shifted embedding.}
    \label{fig:c}
\end{figure*}

We use optical flow as our self-supervised signal and analyze the embedding trained from such flow supervision. Using the same images presented in Figure 2, we visualize the learned embedding in Figure 3. We learned embedding from two sources of optical flow, flow based on \cite{Liu2009} and flow based on PWC-Net pretrained on the Flying Chair dataset, respectively. Two sets of results are shown in parallel in Figure 3. For each flow supervision, from left to right, we first randomly project the learned 512*512*64 embedding map into a 512*512*3 color image in column 1. We also visualize the learned embedding similarity by choosing an anchor embedding vector, as shown by a red plus sign in the image, and calculate the cosine similarity between all other embedding vectors and the anchor vector as shown in column 2. We expect vectors from the same class to have higher similarity scores and thus brighter colors in the map. The flow similarity calculated by Gaussian RBF kernel is also shown in column 3 with sigma set at 0.5. A small region close to the anchor point is bright since the flow in this region are more similar. Since our metric loss is defined by encouraging flow similarity and embedding similarity to be similar across all neighboring elements and the anchor element, we also visualize the neighboring element flow and embedding similarity. To do so, we spatially shift the flow map and embedding map by an offset and then calculate the similarities between the original map and the shifted map using Equations (4) and (6). As illustrated in Figure 3, we shift the flow and embedding map by (5,5) offset in the bottom right direction. The resulting similarity map represents the similarity between any (i, j) element and its neighbor (i+5, j+5) element, where (i, j) and (i+5, j+5) are all within the image dimension. Such defined flow similarity and embedding similarity are shown in column 4 and 5 respectively, where bright color show up in the blood vessel and background, (e.g., both (i, j) and (i+5, j+5) are in the same area) indicating high similarity scores, while dark color occurs at the boundary between vessel and background, (e.g., (i, j) lies on the vessel while (i+5, j+5) lies on the background), indicating that the blood vessel and background have low similarity scores.

Based on only flow supervision, the learned embedding of similar objects is already grouped together as shown by the similar colors in random projection as well as the highlighted area in cosine embedding similarity. The embedding trained with flow based on \cite{Liu2009} captures the overall blood vessel shape, but fails to reveal details and sharp edges. This is mainly due to the fact that the flow used in training appears in vague patches and dilated as illustrated in Figure 2. Therefore the provided supervision signal is not discriminative enough for learning the differences around the vessel boundaries. A lack of discriminative features can be obviously seen in the fluffy similarity map in column 2 and the lack of sharp edges in column 5. On the other hand, embedding trained with PWC-Net flow appears to have successfully captured the semantics as the blood vessel area has a distinct color from the background in both random projection image and embedding similarity map. It is also interesting to note that, although the loss function is defined to encourage similarity between flow and embedding, the learned embedding also picks up some appearance features from seeing large amount of data. For example, for data D2 in the case of PWC-Net flow supervision, the embedding similarity in column 5 shows vessel boundaries that are not present in flow similarity in column 4.

\begin{figure}[htb]
    \center{\includegraphics[width=1.0\linewidth]{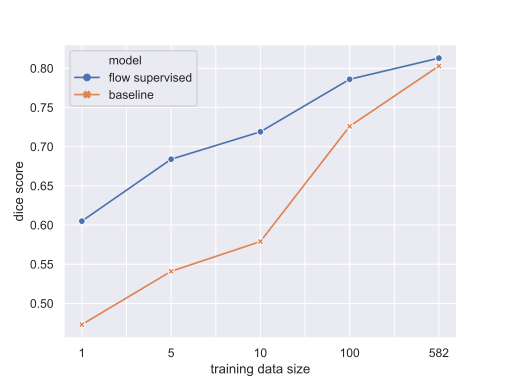}}
    \caption{Dice score of baseline U-Net method and self-supervised method over a range of training data sizes}
    \label{fig:d}
\end{figure}

After learning meaningful pixel embeddings, we then add the last fully convolutional layer of U-Net and finetune the embedding model and last layer together with respectively 1, 5, 10, 100 and 582 annotated training labels, as did in the baseline method for a direct comparison. As can be seen from Figure 4, the dice score of model prediction with flow supervision is consistently better than that without flow supervision across all training data sizes, especially when the number of training data is small. When there is only 1 annotated training data, baseline U-Net training overfits and result in best dice score of 0.473 on the testing set, while for the flow supervised U-Net model, with 1 annotated data, dice score can be improved to 0.605, even better than the baseline model with more than 10 training data. Flow supervised model with 10 or 100 annotated training data performs similarly well compared with baseline U-Net model with 100 or 582 annotated training data respectively. With self-supervision, the demand of annotated data can be dramatically reduced.

\subsection{Ablation experiments}
Observing that sharper edges in the flow map might help with better feature learning, we further finetune the PWC-Net model on X-ray angiography images. We create synthetic X-ray angiography images in a similar fashion as the Flying Chair dataset. We apply random translation and rotation on segmented blood vessels and then alpha blend them with background images to create synthetic data. The background images are extracted from the first few frames of each image sequence when the contrast medium is not yet injected and hence contains only background. The ground truth optical flow is obtained analytically from the synthetic motion transform for both background and vessel. We randomly choose a small amount of segmented data from training set (i.e., 10 data in our case) and randomly transform them to create a large dataset with 5524 pairs of synthetic data consisting of frame 1, frame 2 and their corresponding optical flow map as shown in Figure 5. Despite the small number of real data and simplistic approach used in creating synthetic data, the synthetic dataset is shown to be very useful in finetuning PWC-Net. As shown in Figure 6, compared with the pretrained PWC-Net generated flow shown in column 1, the finetuned model produces much more accurate optical flow, shown in column 2, that captures sharp boundaries and motions of even very small vessel branches. To further verify the correctness of the flow map, we also obtained the reverse flow, shown in column 3, by using the same model and changing the input order of two frames. We use the reverse flow map to warp frame 2 and find the warped results, shown in column 4, compare favorably well with frame 1 as shown in Figure 2. The discrepancy of forward and reverse flow are considered occlusion and shown as dark color in warped images.

\begin{figure}[htb]
    \center{\includegraphics[width=1.0\linewidth]{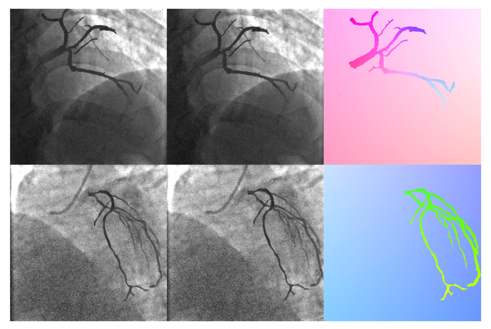}}
    \caption{Sample synthetic X-ray angiography images for finetuning PWC-Net. Two sample data are shown are two rows. On each row from left to right, synthetic frame 1, synthetic frame 2 and their corresponding flow as defined by the transformation}
    \label{fig:e}
\end{figure}

\begin{figure}[htb]
    \center{\includegraphics[width=1.0\linewidth]{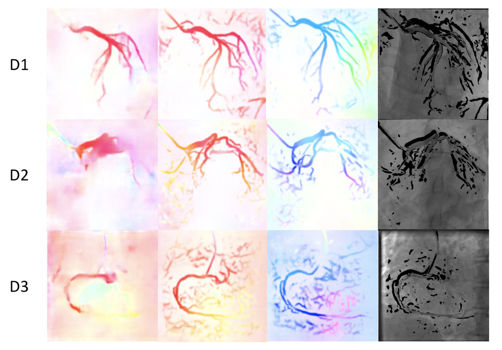}}
    \caption{Optical flow of three sample data on three rows. For each row from left to right, column 1 shows predicted flow by PWC-Net trained on the Flying Chair data, column2 shows predicted flow by PWC-Net trained on the Flying Chair data and finetuned on 10 X-ray angiography synthetic data, column 3 shows the predicted reversed flow by using the same model as in column 2, column 4 shows the warped blood vessel images with occlusion}
    \label{fig:f}
\end{figure}

In the setting of few shot learning where we train on only 1, 5 and 10 annotated data, we compare the baseline model and various embedding model trained with different sources of optical flow, such as flow based on \cite{Liu2009}, PWC-Net flow trained on flying chair dataset, PWC-Net flow trained on flying chair dataset and finetuned by X-ray angiography synthetic dataset, respectively. As shown in Figure 7, the performance with flow based self-supervision are significantly better than the baseline model without any self-supervision. Within the flow supervised candidates, better quality of optical flow also improves the final segmentation performance.

\begin{figure}[!htbp]
    \center{\includegraphics[width=1.0\linewidth]{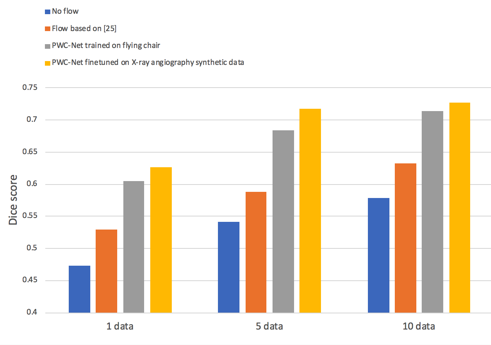}}
    \caption{Performance comparison between model trained without flow supervision and with varying quality flow as supervision. Comparison is made over 1, 5 and 10 annotated training data.}
    \label{fig:g}
\end{figure}

\section{Discussion}
A critical challenge in training deep models arises mostly from the limited number of training samples, compared to the number of learnable parameters. It is especially true in medical field where large amount of annotation is hard to obtain. In this paper, we leverage motion information that is freely available in the raw image sequences to learn meaningful pixel embedding in a self-supervised fashion. We demonstrate that with self-supervision, the same segmentation performance can be achieved with significantly less annotated data while previously can only be achieved with large amount of expensive annotation.
{\small
\bibliographystyle{unsrt}
\bibliography{cvpr}

\begin{thebibliography}{10}

\bibitem{Krizhevsky2012}
Alex Krizhevsky, Ilya Sutskever, and Geoffrey~E Hinton.
\newblock {Imagenet classification with deep convolutional neural networks}.
\newblock In {\em Advances in neural information processing systems}, pages
  1097--1105, 2012.

\bibitem{He2016}
Kaiming He, Xiangyu Zhang, Shaoqing Ren, and Jian Sun.
\newblock {Deep residual learning for image recognition}.
\newblock In {\em Proceedings of the IEEE conference on computer vision and
  pattern recognition}, pages 770--778, 2016.

\bibitem{Ronneberger2015}
Olaf Ronneberger, Philipp Fischer, and Thomas Brox.
\newblock {U-net: Convolutional networks for biomedical image segmentation}.
\newblock In {\em International Conference on Medical image computing and
  computer-assisted intervention}, pages 234--241. Springer, 2015.

\bibitem{Lotter2016}
William Lotter, Gabriel Kreiman, and David Cox.
\newblock {Deep predictive coding networks for video prediction and
  unsupervised learning}.
\newblock {\em arXiv preprint arXiv:1605.08104}, 2016.

\bibitem{Doersch2015}
Carl Doersch, Abhinav Gupta, and Alexei~A Efros.
\newblock {Unsupervised visual representation learning by context prediction}.
\newblock In {\em Proceedings of the IEEE International Conference on Computer
  Vision}, pages 1422--1430, 2015.

\bibitem{Goodfellow2014}
Ian Goodfellow, Jean Pouget-Abadie, Mehdi Mirza, Bing Xu, David Warde-Farley,
  Sherjil Ozair, Aaron Courville, and Yoshua Bengio.
\newblock {Generative adversarial nets}.
\newblock In {\em Advances in neural information processing systems}, pages
  2672--2680, 2014.

\bibitem{Larsson2017}
Gustav Larsson, Michael Maire, and Gregory Shakhnarovich.
\newblock {Colorization as a proxy task for visual understanding}.
\newblock In {\em CVPR}, volume~2, page~7, 2017.

\bibitem{Dosovitskiy2015}
Alexey Dosovitskiy, Philipp Fischer, Eddy Ilg, Philip Hausser, Caner Hazirbas,
  Vladimir Golkov, Patrick {Van Der Smagt}, Daniel Cremers, and Thomas Brox.
\newblock {Flownet: Learning optical flow with convolutional networks}.
\newblock In {\em Proceedings of the IEEE International Conference on Computer
  Vision}, pages 2758--2766, 2015.

\bibitem{Sun2018}
Deqing Sun, Xiaodong Yang, Ming-Yu Liu, and Jan Kautz.
\newblock {Pwc-net: Cnns for optical flow using pyramid, warping, and cost
  volume}.
\newblock In {\em Proceedings of the IEEE Conference on Computer Vision and
  Pattern Recognition}, pages 8934--8943, 2018.

\bibitem{Hui2018}
Tak-Wai Hui, Xiaoou Tang, and Chen~Change Loy.
\newblock {LiteFlowNet: A Lightweight Convolutional Neural Network for Optical
  Flow Estimation}.
\newblock In {\em Proceedings of the IEEE Conference on Computer Vision and
  Pattern Recognition}, pages 8981--8989, 2018.

\bibitem{DeTone2017}
Daniel DeTone, Tomasz Malisiewicz, and Andrew Rabinovich.
\newblock {SuperPoint: Self-Supervised Interest Point Detection and
  Description}.
\newblock {\em arXiv preprint arXiv:1712.07629}, 2017.

\bibitem{Choy2016}
Christopher~B Choy, JunYoung Gwak, Silvio Savarese, and Manmohan Chandraker.
\newblock {Universal correspondence network}.
\newblock In {\em Advances in Neural Information Processing Systems}, pages
  2414--2422, 2016.

\bibitem{Horn1981}
Berthold K~P Horn and Brian~G Schunck.
\newblock {Determining optical flow}.
\newblock {\em Artificial intelligence}, 17(1-3):185--203, 1981.

\bibitem{Stein2007}
Andrew Stein, Derek Hoiem, and Martial Hebert.
\newblock {Learning to find object boundaries using motion cues}.
\newblock In {\em Computer Vision, 2007. ICCV 2007. IEEE 11th International
  Conference on}, pages 1--8. IEEE, 2007.

\bibitem{Jain2017}
Suyog~Dutt Jain, Bo~Xiong, and Kristen Grauman.
\newblock {Fusionseg: Learning to combine motion and appearance for fully
  automatic segmention of generic objects in videos}.
\newblock In {\em Proc. CVPR}, volume~1, 2017.

\bibitem{Cheng2017}
Jingchun Cheng, Yi-Hsuan Tsai, Shengjin Wang, and Ming-Hsuan Yang.
\newblock {Segflow: Joint learning for video object segmentation and optical
  flow}.
\newblock In {\em Computer Vision (ICCV), 2017 IEEE International Conference
  on}, pages 686--695. IEEE, 2017.

\bibitem{Schroff2015}
Florian Schroff, Dmitry Kalenichenko, and James Philbin.
\newblock {Facenet: A unified embedding for face recognition and clustering}.
\newblock In {\em Proceedings of the IEEE conference on computer vision and
  pattern recognition}, pages 815--823, 2015.

\bibitem{Taigman2014}
Yaniv Taigman, Ming Yang, Marc'Aurelio Ranzato, and Lior Wolf.
\newblock {Deepface: Closing the gap to human-level performance in face
  verification}.
\newblock In {\em Proceedings of the IEEE conference on computer vision and
  pattern recognition}, pages 1701--1708, 2014.

\bibitem{Chopra2005}
Sumit Chopra, Raia Hadsell, and Yann LeCun.
\newblock {Learning a similarity metric discriminatively, with application to
  face verification}.
\newblock In {\em Computer Vision and Pattern Recognition, 2005. CVPR 2005.
  IEEE Computer Society Conference on}, volume~1, pages 539--546. IEEE, 2005.

\bibitem{Fathi2017}
Alireza Fathi, Zbigniew Wojna, Vivek Rathod, Peng Wang, Hyun~Oh Song, Sergio
  Guadarrama, and Kevin~P Murphy.
\newblock {Semantic instance segmentation via deep metric learning}.
\newblock {\em arXiv preprint arXiv:1703.10277}, 2017.

\bibitem{DeBrabandere2017}
Bert {De Brabandere}, Davy Neven, and Luc {Van Gool}.
\newblock {Semantic instance segmentation with a discriminative loss function}.
\newblock {\em arXiv preprint arXiv:1708.02551}, 2017.

\bibitem{Ross2009}
Michael~G Ross and Leslie~Pack Kaelbling.
\newblock {Segmentation according to natural examples: Learning static
  segmentation from motion segmentation}.
\newblock {\em IEEE transactions on pattern analysis and machine intelligence},
  31(4):661--676, 2009.

\bibitem{Pathak2017}
Deepak Pathak, Ross~B Girshick, Piotr Doll{\'{a}}r, Trevor Darrell, and Bharath
  Hariharan.
\newblock {Learning Features by Watching Objects Move.}
\newblock In {\em CVPR}, volume~1, page~7, 2017.

\bibitem{Mahendran2018}
Aravindh Mahendran, James Thewlis, and Andrea Vedaldi.
\newblock {Cross Pixel Optical Flow Similarity for Self-Supervised Learning}.
\newblock {\em arXiv preprint arXiv:1807.05636}, 2018.

\bibitem{Liu2009}
Ce~Liu.
\newblock {Beyond pixels: exploring new representations and applications for
  motion analysis}, 2009.

\bibitem{Kirbas2004}
Cemil Kirbas and Francis Quek.
\newblock {A review of vessel extraction techniques and algorithms}.
\newblock {\em ACM Computing Surveys (CSUR)}, 36(2):81--121, 2004.

\bibitem{Frangi1998}
Alejandro~F Frangi, Wiro~J Niessen, Koen~L Vincken, and Max~A Viergever.
\newblock {Multiscale vessel enhancement filtering}.
\newblock In {\em International Conference on Medical Image Computing and
  Computer-Assisted Intervention}, pages 130--137. Springer, 1998.

\end{thebibliography}
}
\end{document}